# Perspectives for Strong Artificial Life

Jean-Philippe Rennard
Grenoble Ecole de Management
jp@rennard.org

**Abstract**: This text introduces the twin deadlocks of strong artificial life. Conceptualization of life is a deadlock both because of the existence of a continuum between the inert and the living, and because we only know one instance of life. Computationalism is a second deadlock since it remains a matter of faith. Nevertheless, artificial life realizations quickly progress and recent constructions embed an always growing set of the intuitive properties of life. This growing gap between theory and realizations should sooner or later crystallize in some kind of "paradigm shift" and then give clues to break the twin deadlocks.

## 1 Introduction

"The ultimate goal of the study of artificial life would be to create 'life' in some other medium, ideally a *virtual* medium where the essence of life has been abstracted from the details of its implementation in *any* particular hardware. We would like to build models that are so life-like that they cease to be *models* of life and become *examples* of life themselves." (Langton, 1986, p. 147, original emphasis).

This statement launched a large debate among biologists, philosophers and computer scientists. It defined what is now known as *strong artificial life* i.e. the fact that artificial life is not limited to theoretical biology or biomimetic artefacts, but can be extended to the creation of new instances of life, possibly independent of any material medium. With the emergence of artificial life as a scientific field, the question of the creation of life by men therefore left the fields of religion or superstition to join the scope of science.

Nearly two decades later, the debate about strong artificial life remains very active. Despite remarkable realizations — such as *Venus* (Rasmussen, 1990), *Tierra* (Ray, 1992) or *Cosmos* (Taylor, 1999a) — and pompous declarations such as the famous *How I Created Life in a Virtual Universe* (Ray, 1993), we still have no strong demonstration of the feasibility of strong artificial life.

The problem was nicely introduced by a provocative reasoning presented by S. Rasmussen at the second artificial life workshop (Rasmussen, 1992):

I. A universal computer at the Turing machine level can simulate any physical process (Physical Church-Turing thesis).

II. Life is a physical process (von Neumann).
III. There exist criteria by which we are able to distinguish living from non-living objects.

Accepting (I), (II) and (III) implies the possibility of life in a computer.

IV. An artificial organism must perceive a reality $R_2$, which for it is just as real as our "real" reality $R_1$ is for us.
V. $R_1$ and $R_2$ have the same ontological status.
VI. It is possible to learn something about fundamental properties of realities in general, and of $R_1$ in particular, by studying the details of different $R_2$'s. An example of such a property is the physics of a reality.

Even though points (IV) to (VI) tend to justify the huge potential contribution of "artificial realizations" to science, yet, point (I) (computationalism) on the one hand and points (II) and (III) (concept of life) on the other hand found what can be called the *twin deadlocks of strong artificial life*.

This chapter will first present both deadlocks and show the current limits of purely theoretical approaches. It will then present some recent realizations and show the growing gap between the relative stagnation of theory and the progress of experiments. Finally, it will examine the consequences of this hiatus.

**2 On the Concept of Life**

Life is obvious; a five years old boy can easily distinguish an inert object from a living one. Nevertheless, thousands pages have been written since Aristotle about the concept of life, and the problem remains clearly unsolved. Some definitions are essentially conceptual, they try to propose a strong theoretical basis for life. Because of the difficulty to define a unifying concept, other definitions are mainly empirical and based on the observation and the research of apparent invariants.

**2.1 Limits of conceptual definitions of life**

According to Dawkins, the word "living" does not necessarily refer to something definite in the real world (Dawkins, 1996, p. 39). For him, the systems we call "living" emerged from a cumulative and very progressive process favoring the most efficient replicators i.e. the replicators which had the best ability to embed a "surviving machine". That is therefore the same processes which were at work from the formation of the first "soap bubbles" to the emergence of robust replicators. This suggests a continuity between living and non-living. It is then not possible to define a clear borderline dividing the inert on the one hand and the living on the other hand.

There is now a growing consensus among biologists concerning this continuity : "It seems more appropriate to consider life as a continuum property of organizational patterns, with some more or less alive than others." (Farmer & Belin, 1992, p. 819). "Life is not an absolute quality which suddenly appeared, it gradually emerged at the beginning of the evolution. […] there is no clear boundary line." (Cairns-Smith, 1990, p. 14). "Presumably, life emerged in a gradual, long-term process whereby primitive proto-organisms were created in the teeming, primal soup of the seas. Life is still not completed; it is a partly indefinite phenomenon of becoming, not a completely determined state of being." (Emmeche, 1994, p. 37).

The fact that all the dominant modern theories consider that life progressively emerged from some kind of proto-organisms, strongly grounds the hypothesis of the existence of a large and blurring borderline between living and non-living, were the inert progressively fades into the living. This continuity makes the conceptualization of life a very difficult task, since it should exist "half-living" or "quite-living" systems. Main concepts of life fail to manage this continuum. Let us see some significant examples.

### 2.1.1 Life: dissipative structures

Famous physicist E. Schrödinger first proposed in 1944 an explicitly thermodynamic definition of life: "What is the characteristic feature of life? When is a piece of matter said to be alive? When it goes on 'doing something', moving, exchanging material with its environment (…) It is by avoiding the rapid decay into the inert state of 'equilibrium' that an organism appears so enigmatic (…) How does the living organism avoid decay? The obvious answer is: By eating, drinking, breathing and (in the case of plants) assimilating. The technical term is *metabolism*." (Schrödinger, 1986, pp. 168-171, original emphasis).

For Schrödinger, life is foremost the capacity for a material system to maintain its thermodynamic disequilibrium toward the environment. According to the Nobel Prize Ilya Prigogine (Nicolis & Prigogine, 1977), such systems are conceptualized as "dissipative structures", i.e. structures in which non linear interactions, while consuming energy and through phase transitions, are driven to a more complex, unpredictable organization.

It is now largely admitted that living systems are dissipative structures; that by using energy they are able to create order out of chaos, but this property is also known to be insufficient to define life. Many dissipative structures are clearly inert (candle flames, vortexes…).

Conceptualization of dissipative structures strongly relies on "bifurcations", "symmetry breaks" or "phase transitions". The evolution toward dissipative structures is then very sudden; there is no smooth evolution but catastrophic transformations. This concept therefore clearly violates the continuity of the transition from inert to living.

### 2.1.2 Life : autopoietic systems

Looking for a definition of life in cognitive science, H. Maturana and F. Varela proposed a more sophisticated approach based upon autonomy and self-reference (Maturana & Varela, 1974). They conceptualized life as an *autopoietic system*:

"*An autopoietic system is organized as a network of processes of production of components which: (a) through their transformations and interactions continuously regenerate the network of processes that produced them, and (b) constitute the system as a concrete unity in the space in which it exists by* *specifying the topological domain of its realization as a network.*" (Varela, 1989, p. 45, original emphasis).

Self-reference and operational closure are interpreted as the root properties of the construction of an *identity* which is dynamically self-preserved, i.e. which is able to react to environmental perturbations in such a way that it can preserve its own internal equilibrium (autopoiesis can then be interpreted as an extension of homeostasis). Autopoiesis is what distinguishes living systems from inert ones: "Autopoiesis in the physical space is a necessary and sufficient condition for a system to be a living one." (Maturana & Varela, 1980, p. 84).

The reference to the physical space means that living systems are necessarily metabolic systems, and allows to distinguish them from systems which can be seen as autopoietic but not alive (societies for example —see Boden, 2000).

The autopoietic definition of life is very elegant. It conceptualizes life with a unique and fundamental concept, which seems to be accurate and adequately adapted to observation. But once more, this definition is insufficient. Some physical phenomenon, like Bénard cells, are autopoietic systems in the physical space, but are nevertheless clearly inert. It has been argued that this problem could be overcome by adding a simple condition: living systems are autopoietic systems in the physical space which embed a self-description (Heudin, 1994). This addition is useful but *ad hoc*.

Because of operational closure, autopoiesis also violates the continuity property of life. A system is operationally closed or not, half-closure is impossible. Therefore autopoiesis is an either/or property, a system can't be semiautopoietic.

**2.1.3 Life: adaptive systems**

In order to propose a unifying concept of life, some authors (e.g. Cairns-Smith, 1990; Maynard Smith & Szathmáry, 1995) emphasized the evolutionary property of the living. Thus, according to E. Mayr: "We shall regard as alive any population of entities which has the properties of multiplication, heredity and variation." (Mayr, 1975, pp. 96-97). In the field of artificial life, the most famous evolutionary definition is the "supple adaptation" concept of M. Bedau (Bedau, 1996; Bedau, 1998a). Supple adaptation refers to the fact that living systems have an unending capacity to find new unpredictable adaptive solutions to unanticipated changes in the environment.

"(…) the entity that is living in the *primary* sense of that term is the supplely adapting system itself. Other entities that are living are living in a *secondary* sense by virtue of bearing an appropriate relationship to a supplely adapting system." (Bedau, 1996, p. 339, original emphasis).

That is therefore the whole population which is "primary" living. The individuals are living only because of their relationship to the supplely adapting system constituted by the population.

Such a definition is highly counter-intuitive, since it is very difficult to accept life as a population property. Moreover, Bedau does not really answer the problem of the eventuality of long periods of evolutive stability (stasis). Despite this stability, the individuals living there would clearly be alive.

This is nevertheless one of the most attractive modern definitions. It is both concise and able to manage problematic cases such as mules or viruses, and even spores or frozen sperm (which are dormant and then temporarily not connected to the living system they belong to).

According to Bedau (Bedau, 1998a), the supple adaptation definition conforms to the continuity hypothesis. The same way life is a matter of degree, supple adaptation can vary to different degrees. But even though the supple adaptation capabilities can grow or decrease, there is nevertheless a clear distinction between adaptive systems and not-adaptive ones, even when this adaptability is close to zero. We now have many proofs of the unity of life (e.g. the universality of the genetic code, the chirality of biological molecules…). Therefore, the oldest self-replicating structures life descends from were already (potentially) supply adapting structures.

These examples —which are representative of important classes of conceptual definitions of life— show how difficult it is to manage the continuity from nonlife to life, and the growing gap between life theories and what biology thinks life originates from. It has been argued that this continuity is not a real problem, since, for example, the continuity from green to blue doesn't prevent the recognition of both colors. In fact, this doesn't prove anything, since neither green, nor blue really exist, both colors have been arbitrarily defined, the same way the A note in music has been defined to be a 440 Hz sound.

This doesn't mean that a conceptual definition of life is impossible, even though such a conceptualization will inevitably embed some arbitrarily set borderline (or "borderzone"), but we still lack a concept able to cover the whole range from non living to living. Continuity remains one of the main obstacles on the way toward a theory of life.

**2.2 Cluster definitions of life**

Considering the difficulty of a conceptual definition of life, some researchers have tried to propose an empirical approach based on observation and the identification of invariants. The definitions used by the general public are of this type. Thus, according to Merriam-Webster dictionary, life is:

- The state of a material complex or individual characterized by the capacity to perform certain functional activities including metabolism, growth, reproduction, and some form of responsiveness or adaptability.

Life is seen as the result of a complex material organization which performs some specific tasks. Such a definition doesn't say anything about this organization, it only considers life as a cluster of properties (growth, self-reproduction…). Such definitions are common. In the field of artificial life, the most famous cluster definition is that of Farmer and Belin who selected eight criteria (Farmer & Belin, 1992, p. 818):

1. Life is a pattern in spacetime.
2. Self-reproduction.
3. Information storage of a self-representation.
4. Metabolism.
5. Functional interactions with the environment.
6. Interdependence of parts.
7. Stability under perturbations.
8. Ability to evolve.

This definition is often used because of its efficiency. It is difficult to remove a property and (usually) not necessary to add a new one. It nevertheless poses two main problems:

- It is not sufficient to identify all forms of life. For example, it doesn't efficiently qualify viruses or the proto-organisms that are supposed to be at the origin of life.
- It doesn't say anything about what life is, it only notices an empirical and intuitive set of properties without giving any explanation of their origins.

All the definitions of life based on a set of properties share the second problem. Being essentially empirical, they are unable to explain why they have to consider a given property.

The issue is how to select meaningful properties? For example, why to include self-reproduction, or evolvability? Both these properties can be seen as qualifying life, but they only have been considered relatively to the sole example of life we know, nothing proves that they qualify life itself.

This problem has no direct solution since any empirical definition of life is necessarily based on "life as we know it" and not on "life as it could be". This is known as the *Small Sample Problem*: "The *Small Sample Problem* is the problem of determining from a small sample (in this case a sample of one) which features of some things (in this case, life) are essential to that thing." (Cameron, 2000). One of the main purposes of science is to determine fundamental invariants, how to define invariants with only one instance of a phenomenon?

This doesn't mean that defining life is impossible, but that, as far as we only know earth life, it is impossible to base it only on purely observational features, one inevitably needs some general theory or principles to overcome this problem.

More than 2,000 years after Aristotle works, the concept of life then clearly remains to be defined. The continuity problem and the small sample problem make the question quasi insoluble. Because of the small sample problem, it is impossible to strongly ground life invariants. Because of the continuity problem, it is very difficult to find a conceptual definition able to manage the (possibly large) area between the inert and the living (even though it is theoretically possible). This is such a fundamental problem, that the question of the scientific status of life remains open:

"(…) 'life' may not be a scientifically grounded category (such as water, or tiger) whose real properties unify and underlie the similarities observed in all those things we call 'alive'. Instead, it may pick out a rag-bag of items with no fundamental unity." (Boden, 1996, p. 1). Yet, the only way we have to define life, is to find processes or principles which could explain why some specific features are gathered in living systems (complexity, evolvability…): "In the final analysis, the nature of life will be settled by whatever provides the best explanation of the rich range of natural phenomena that seem to characterize living systems." (Bedau, 2003). It means that the so-called "intuitive properties of life" remain of central importance. Computational constructions generating a great deal of these properties could demonstrate some essential principles regarding life.

**3 On Computationalism**

"[Those who know automata] will consider the body as a machine which, being built by God, is infinitely best organized (…) than any machine built by men (…) if there were such machines having the organs and the face of a monkey (…) we couldn't recognize that they are not of the same nature as these animals." (Descartes, 1999, pp. 69-70).

Published in 1637, this sentence shows that computationalism is not a new idea. The form of computationalism proposed by René Descartes is very different from the modern one (notably because of the *dualism*, which considers that the mind has to be interpreted in a different way than the body), but the fundamental idea is already present: *the body is a machine*.

In modern science, this idea is formally based on the so-called physical Church-Turing thesis, which can strongly be interpreted this way: "(…) any physical process can be thought of as a computation, therefore any physical process can be re-created in a computational medium." (Helmreich, 1998, p. 75). On this basis, functionalism (which is an enlarged version of computationalism, not limited to computers) poses the possibility of *multiple-realizability*. Just as there are different ways to realize, for example, means of transport (cars, planes, roller skates…), there are different ways to realize a mind or a living being. The important point is not the physical realization but the properties of the system. The ultimate consequence of this thesis in the field of artificial life is that: "(…) artificial life is possible precisely because living *organisms* themselves are types of *machines* that can reproduce themselves. And since a machine's functions —its logic— can in principle always be imitated in new constructions (whether with the same or other material is unimportant), it follows that life itself, the organic machine can be constructed." (Emmeche, 1994, p. 49, original emphasis). Therefore emerges: "(…) the idea that computers are instances of biological process." (Sober, 1996, p. 362).

The idea that a discrete computing system can realize any physical process has recently been promoted to a "Principle" by Stephen Wolfram with his *Principle of Computational Equivalence* (Wolfram, 2002) which assumes that all non trivial processes, natural or produced by humans, can be viewed as computations, and that those computations are of equivalent sophistication.

Despite apparently strong mathematical and philosophical bases, computationalism (and functionalism) is strongly debated. Some objections are essentially formal. For example, it has never been proved that a digital computing system is equivalent to an analog one. On the contrary, it has been proved that some analog computing devices could have super-Turing capabilities. H. Siegelmann (Siegelmann, 1995) showed for example that analog recurrent neural networks could perform operations beyond the capabilities of Turing machines. More generally, Copeland and Sylvan showed that "(…) the set of well-defined computations is not exhausted by the computations that can be carried out by a Turing machine." (Copeland & Sylvan, 1999, p. 46). But these proofs are controversial since the considered machines seem impossible to be physically realized.

Others objections are more philosophical and based on notions such as representation or semantic, the most famous being the Chinese room experiment of Searle (very briefly: if you have an appropriate set of rules, you can write an answer in chinese to a chinese question without understanding a word of chinese —Searle, 1990). Even though it has been argued that this experiment doesn't apply to artificial life (Anderson & Copeland, 2003), it remains one of the most discussed objection.

The problem of implementation also remains completely open. Pushing the idea of a correspondence between physical states and computation to the limit even tend to empty computationalism: "if (…) it could be shown that for any computational description and any given physical system, one can find 'physical states' of that system that can be set in correspondence with the computational ones and are, furthermore, appropriate (…) then computational explanations would be in danger: every system could then be seen to compute! In other words, computationalism would be vacuous if every physical system could be viewed as implementing every computation." (Scheutz, 2002).

The debate between computationalists and their critics is an unending one; each argument always has its counter-argument. Fundamentally, computationalism essentially remains a matter of faith since, no matter of the precision of a measurement and of the result, detractors of computationalism can always say that a finer measurement would show a continuum and supporters that it would show discrete processes (Sullins, 2001). Therefore, and as long as a deeper concept of the ontological status of computational realizations has not emerged, computationalism can neither be proved, nor be refuted (see Delahaye, 2002).

**4 On Realizations**

The twin deadlocks of strong artificial life don't prevent the multiplication of experiments which present more and more properties of life as we know it.

A decade after its presentation, the most famous experiment remains Tom Ray's Tierra (Ray, 1992), some kind of artificial universe where matter is represented by CPU instructions, energy by CPU time and space by computer memory, in which algorithmic "creatures" mutate and self-reproduce. Ray circumvented the problem of life definition by asserting that to be living, a system only has to be capable of self-replication and open-ended evolution. Open-ended evolution refers to "(…) a system in which components continue to evolve new forms continuously, rather than grinding to a halt when some sort of 'optimal' or stable position is reached." (Taylor, 1999b, p. 34).

Ray's results have been sufficiently impressive to give rise to comments such as: "From a purely logical point of view, the barrier between life and artificial life seemed to have come down: the universality of life was proven." (Adami, 1998, p. 49).

C. Emmeche examined Ray's creatures according to Farmer and Belin's eight criteria (Emmeche, 1994, p. 43-46):

1. Ray's creatures are information structures rather than material objects.
2. They are able to self-reproduce.
3. They have self-representation.
4. They have some kind of metabolism since they redistribute some of the computer's electrical energy.
5. They have functional interactions with their environment.
6. Their components are mutually interdependent and they can die.
7. They are stable in their environment.
8. They can evolve.

According to Emmeche, only the properties 2 (self-reproduction is essentially formal, it doesn't consume any "matter"), 4 (is it reasonable to consider alterations of electromagnetic states as a metabolism?) and 7 (the considered stability is very weak) are not fully satisfied.

The question here is not to say that Ray's creatures are "quite living", but to point out that such a construction satisfies many of the intuitive properties of life.

Despite strong criticism, Tierra inspired many works, the most accomplished ones being Avida (Adami, 1998) or Cosmos (Taylor, 1999a). Tierra-like "creatures" are most probably one of the closest from life realizations. Nevertheless it is very difficult to consider them as living. Tierra's creatures even don't comply Ray's definition of life. Actually, it has been shown that these constructions don't really support "open-ended" evolution. Using some "evolutionary activity" indicators, Bedau *et al.* (Bedau, Snyder, Titus Brown, & Packard, 1997) showed that the evolutionary process in these constructions is quantitatively and qualitatively different from the biosphere. In the real world, the evolutionary process seems to never stagnate. Biosphere always proposes new environments, new niches and allows growing adaptive successes. On the contrary, with Evita (a Tierra-like —see Bedau & Titus Brown, 1999) or Bugs (an animat-like —see Packard, 1989), after a phase of growth, adaptive success quickly stagnates and remains constant. Similar results have been obtained (Ward, 2000) with Tierra and Echo (Holland, 1992).

Two main problems are generally pointed out to explain the inability of Tierra-like constructions to generate an unending evolutionary process: (1) the environment is both small and insufficiently diversified; (2) the "biochemical" laws are set by the creator and are quite simple and fixed. Unlike in the real world, there is no possibility of selection of an efficient set of chemical processes eventually able to support open-ended evolution.

Some recent constructions propose ways to overcome these limits. The available space can be greatly enlarged by using Internet; that was the idea of Tom Ray when he proposed NetTierra (Ray, 1995) which was first implemented in 1997 between universities all around the world. Ray thinks that like in Amazonia where the diversity relies more on the multiplication of interactions between species than on the specificity of the ecological substrate, the complexity of the digital organisms and of their interactions in NetTierra could lead to a self-sustained evolutionary dynamic (Ray, 2001).

The problem of the construction of an endosemantic evolutionary environment is more complex. Pargellis recently proposed Amoeba-II (Pargellis, 2001), an environment designed to test the "spontaneous" emergence of self-replicating "creatures". Unlike Amoeba-I (Pargellis, 1996), the assembler of Amoeba-II is Turing-universal and allows the system to partially define its own genetic code. With this work, Pargellis is close to the growing field of *Artificial Chemistry*, which tries to build systems similar to the real chemical system, and to analyze the origin of evolutionary units (Dittrich, 2001).

Yet, artificial chemistries, despite great success in various fields, failed to produce an open-ended evolutionary system, but recent developments tend to show that it could be possible. New approaches try to embed artificial creatures in their environment i.e. try not to differentiate the constituents of the environment and of the creatures (one notices that this also is one of the directions suggested by the "next generation" of computationalism —see Sheutz 2002). Therefore important properties (e.g. self-replication of the genome) are implicitly encoded. "[To realize open-ended evolutionary systems] no representational distinction should exist between phenotypes and the abiotic environment. Rather, the important representational distinction is between genotypes (viewed as relatively inert, symbolic structures) and phenotypes plus abiotic environment (a dynamical system)." (Taylor, 2002b, p. 12). This approach tries to realize the "semantic closure" of Pattee: "(…) self-reference that has open-ended evolutionary potential is an autonomous closure between the dynamics (physical laws) of the material aspects and the constraints (syntactic rules) of the symbolic aspects of a physical organization. I have called this self-referent relation semantic closure (…)." (Pattee, 1995, see also Rocha, 2001). Such a construction could lead to "creative evolution" (Taylor, 2002c), allowing complex mechanisms such as exaptation (modification of the use of a character) to occur. First results have been obtained with EvoCA (Taylor, 2002b), where the environment is represented by a layer made of a cellular automaton (physical laws, then dynamic) and the genotypes by a specific second layer (syntactic rules, then inert). Each genotype controls a given cell in the first layer (also using preconditions specified by the neighborhood of the cell) and —in the first version— evolves through a genetic algorithm (figure 1).

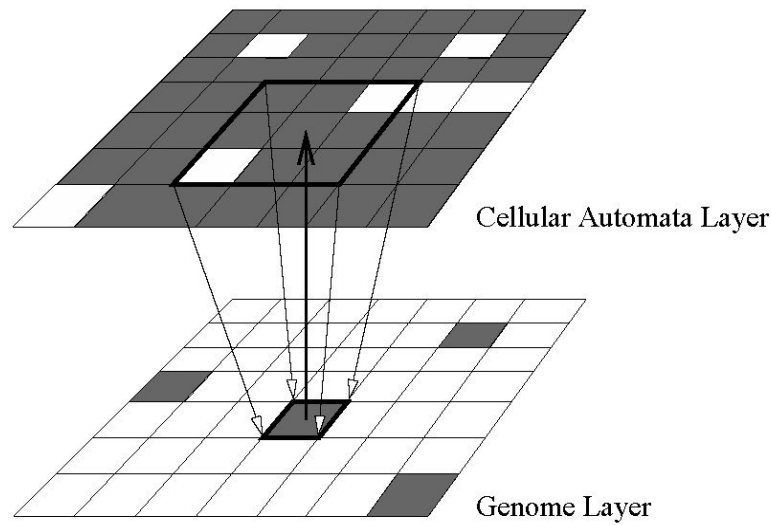

Figure 1: Layers structure of EvoCA. The figure shows how a genome controls a particular cell, using conditional genes linked to the neighborhood. The position of the controlled cell is relative to that of the genotype. (Taylor, 2002a).

Yet, EvoCA is still in its infancy. Taylor only demonstrated the capacity of its construction to support an evolutionary process. According to a given goal (a specific configuration of the cellular automaton) EvoCA has evolved a structure with a more than 80% fitness (figure 2).

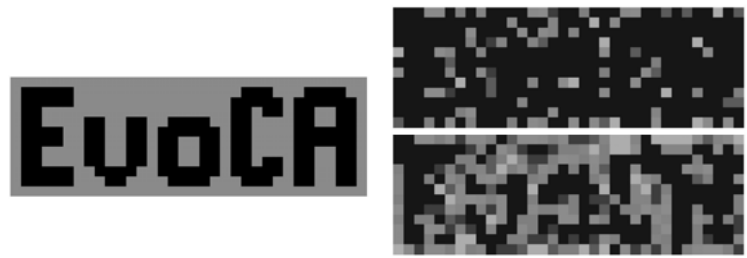

Figure 2: Goal, initial configuration and final result where one can distinguish the letters EvoCA (non-quiescent cells are lights) (Taylor, 2002b).

The next step (EvoCA-B) supposes the suppression of the external goals and the use of endogenous selection. Individuals will have to survive and to reproduce (Taylor, 2002b). Syntactic rules will have to construct the right compromise with the dynamic of the cellular automaton. By defining fitness through the dynamic of the cellular automaton —i.e. by suppressing any externally defined goal— the system will try to approach semantic closure. According to Taylor, by succeeding, these works could lead to systems capable of supporting autopoietic, self-replicating structures. EvoCA-B still has to prove its capabilities and the system remains far from a really endogenous one. Nevertheless, this is a promising attempt to unify phenotypes and the abiotic environment, which is an essential prerequisite on the way toward autopoietic structures capable of an open-ended evolutionary dynamic.

Important progresses have also been made in the construction of multi-cellular, self-replicating systems. D. Mange *et al.* recently proposed the *Tom Thumb algorithm* "[that] will make it possible to design a self-replicating loop with universal construction and universal computation that can easily be implemented into silicon." (Mange, Stauffer, Petraglio, & Tempesti, 2003).

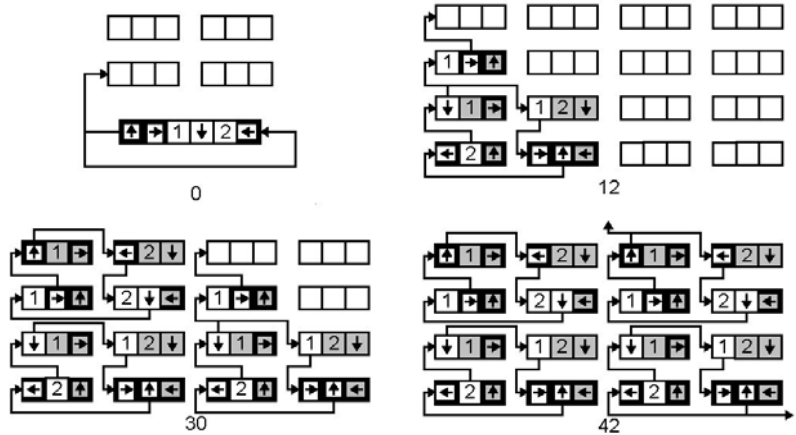

Figure 3: The minimum cell is made of four molecules (2×2) storing three hexadecimal values, and the original genome is a string of six hexadecimal values (step 0). One sees the first cell (step 12) and the construction of 3 daughters cells. (Mange, Stauffer, Petraglio, & Tempesti, 2003).

Mange *et al.* demonstrated that the Tom Thumb algorithm is able of universal construction and computation. It means that it is theoretically possible to realize a self-replicating loop of an arbitrary complexity. Such a construction could notably be of very high interest in the field of nanotechnologies where self-replication is of central importance (see for example Drexler, 1987; Roco, 2002).

Implemented in FPGA (Biowall, see Tempesti, Mange, Stauffer, & Teuscher, 2002), the results are so impressive that Mange *et al.* tried to test its conformance to Varela, Maturana and Uribe autopoiesis checklist (Varela, Maturana, & Uribe, 1974):

- The cell has identifiable boundaries.
- It is made of a set of parts.
- The functioning depends on complex interactions between components.
- The boundaries result from mutual interactions between neighborhood constituents.
- That is the same process that produces boundaries and the cell itself.
- All constituents are produced by other constituents.

"Our self-replicating loop with universal construction is thus an autopoietic cell in the space within its molecule exists. According to Varela *et al.*, such an autopoietic cell has the phenomenology of a living system." (Mange, Stauffer, Petraglio, & Tempesti, 2003). Mange *et al.* of course don't consider this construction as living, but they show that this construction embed some very specific properties of living systems, particularly a sort of autopoietic dynamic.

Both EvoCA and Tom Thumb algorithm are only illustrative examples, but they clearly show that recent developments lead to constructions embedding an always growing set of the intuitive properties of life. The universality of Tom Thumb algorithm could lead to self-replicating constructions of an arbitrary complexity. Ultimately, descendants of EvoCA-like constructions could lead to operationally closed evolutive structures, intimately embedded in a dynamic environment (then having some metabolic capacities), capable of self-replication and self-maintenance.

Recently, S. Rasmussen *et al.* presented their works on proto-organisms (Rasmussen, Chen, Nilsson, & Abe, 2003). They consider as living proto-organisms having the ability to evolve, to self-reproduce, to metabolize, to respond adaptively to environmental changes, and to die. Once again, we are far from the objective, but, *because of their materiality*, anybody which would examine such proto-organisms would undoubtedly consider them as living. Even though they won't be based on DNA, no essential property would differentiate this form of artificial life from life as we know it. What about future computational realizations also embedding these properties? According to Olson: "It seems to me that computer-generated organisms are problematic for the same reason as computer-generated mountains are." (Olson, 1995, p. 1). This argument is common but not tenable. A simulation is controlled by its designers. On the contrary, the self-organization dynamic of artificial worlds —e.g. complex transients in class IV cellular automata (Wolfram, 1984)— *emerges* from the local "physical laws". Likewise, the properties of the "potentially living" systems will *emerge* from the interactions between semantic rules and physical laws. The properties characterizing these constructions will be the consequences of the own dynamic of the computational environment, not of the decisions of the designer (who is anyway unable to predict the evolution of its construction). These emergences demonstrate that specific properties such as self-reproduction or evolvability are multiply realizable. The process is then ontologically different from simulations.

One can always argue that even though some computational constructions are not mere simulations, it is possible to consider that a second order reality doesn't have the same ontological status than a first order one. For example, Sullins considers that Rasmussen's points IV and V, we presented above, are circular: "[Rasmussen] is making the claim that an artificial reality created in the computer is able to capture all of the essential qualities of our reality ($R_1$ is equal to $R_2$) as long as living agents are interacting with the system, but the artificial reality must already be ontologically equivalent to our reality in order to produce truly living artificial life forms." (Sullins, 1997). This argument is very interesting, but "Why should one have priority over the other?" (Rasmussen, 1992, p. 770), and what about our reality (which furthermore is not proved to be of first order) were life undoubtedly exists?

In this context, it seems to us that the philosophical basis to consider proto-organisms as living, and "virtual" organisms (i.e. belonging to a so-called second order reality) having the same emergent properties as mere simulations, are quite weak.

**5 Conclusion**

"The borders between the living and the nonliving, between the Nature-made and the human-made appear to be constantly blurring." (Sipper, 2002, p. 222). Nearly twenty years after the official birth of artificial life, and despite a huge amount of very high quality philosophical works, the problem of the realization of strong artificial life remains with no nearer solution. It has been said that artificial life is *de facto* a symptom of the "end of science" (J. Horgan quoted in Sullins, 2001) but, opposite to purely philosophical works, computer simulations never ceased to progress. Progresses in artificial life constructions will necessarily contribute to the evolution of philosophical thought at least in two ways:

- They will help to validate hypothesis and thought experiments. J. Sullins (Sullins, 2001) considers computer simulation as a tool for what he called "the new computer assisted discovery process"; the best example being von Neumann self-replicating cellular automata which demonstrated a process similar to the biological one, discovered some years later. Bedau also greatly emphasizes the capability of artificial life to help the emergence and the validation of philosophical hypothesis: "Artificial life simulations are in effect thought experiments; but *emergent* thought experiments. (…) What is distinctive about emergent thought experiments is that what they reveal can be discerned only by simulation; armchair analysis in these contexts is simply inconclusive. Synthesizing emergent thought experiments with a computer is a new technique that philosophers can adapt from artificial life." (Bedau, 1998b, p. 144, original emphasis). New simulations, associated with new hypothesis on the origin of a given life property (moreover a given set of life properties) will necessarily increase in the coming years. Comparisons between artificial systems and real living ones will inevitably give clues to the comprehension of life.
- The same way the study of cellular automata and their classification showed that complexity can occur only in a specific range of universal laws (Langton, 1990), the multiplication of more and more accurate simulations of life —even in the absence of any new hypothesis— can help the discovery of unknown underlying processes and connections.

In this context, artificial life is a new tool for discovery, the same way the microscope or the telescope were on the 17$^{th}$ century. The growing gap between theory and observations is very common in the history of science and, according to T. Kuhn (Kuhn, 1996) it is one of the main source of paradigm shifts. The same way that improvements in skies observation progressively leaded from pure observation (T. Brahe) to empirical conceptualization (J. Kepler) and then to a general theory (I. Newton), or that progresses in the study of fossils led from Cuvier to Lamarck and Darwin, computer simulations of life provide a unique tool to help the construction of a general theory of life. Sooner or later, some experiments (computational and physical) will be so life-like that the philosophical interpretation of life will inevitably strongly be influenced. This evolution, i.e. the fact that computer simulations will satisfy a growing set of the intuitive properties of life, will inevitably contribute to overcome the twin deadlock of strong artificial life:

- The problem of the capability of a computing system to reproduce any physical process would no longer be a deadlock. In this context, the issue is not to show the validity of computationalism, but only to show that a computing system is able to embed constructions having the properties of living systems. The demonstration of the feasibility of such constructions remains of course to be done, but the experiments we presented show that we are on the way toward it.
- "[The concept of life] is an historical artifact, which varies across different cultures and which changes as our beliefs and preconceptions evolve." (Bedau In Wheeler et al., 2002, p. 88). Computational constructions based on principles able to unify the intuitive properties of life will ease the elaboration of a scientific definition of life, but they should also contribute to the evolution of the cultural interpretation of what life is. By succeeding, this approach would empty the issue of strong artificial life, since life would be redefined (in some direction or other) according to computational results.

The great physicist Andrei Linde recently declared: "We have always used science to improve living conditions. From now on, we must make science a tool to understand life, consciousness and, above all, ourselves…" (Benkirane, 2002, p. 376). Artificial life and attempts to demonstrate the validity (possibly) of its strong interpretation are unprecedented tools to help science evolve toward Linde wishes.

**Acknowledgments**: We particularly thank the reviewers for their constructive remarks.